\documentclass[runningheads]{llncs}

\usepackage{eccv}

\usepackage{multirow}

\usepackage{url}

\definecolor{Gray}{gray}{0.9}

\usepackage{eccvabbrv}

\usepackage{graphicx}
\usepackage{booktabs}

\usepackage[accsupp]{axessibility}  

\usepackage[pagebackref,breaklinks,colorlinks,citecolor=eccvblue]{hyperref}

\usepackage{orcidlink}

\begin{document}

\title{A Large-scale Medical Visual Task \\ Adaptation Benchmark} 

\titlerunning{A Large-scale Medical Visual Task Adaptation Benchmark}

\author{Shentong Mo\inst{1}\thanks{shentongmo@gmail.com} \and
Xufang Luo\inst{2} \and
Yansen Wang\inst{2} \and Dongsheng Li\inst{2}
}
\authorrunning{S.~Mo et al.}

\institute{Carnegie Mellon University \and
Microsoft Research}

\maketitle

\begin{abstract}

Visual task adaptation has been demonstrated to be effective in adapting pre-trained Vision Transformers (ViTs) to general downstream visual tasks using specialized learnable layers or tokens.
However, there is yet a large-scale benchmark to fully explore the effect of visual task adaptation on the realistic and important medical domain, particularly across diverse medical visual modalities, such as color images, X-ray, and CT.
To close this gap, we present Med-VTAB, a large-scale Medical Visual Task Adaptation Benchmark consisting of 1.68 million medical images for diverse organs, modalities, and adaptation approaches. 
Based on Med-VTAB, we explore the scaling law of medical prompt tuning concerning tunable parameters and the generalizability of medical visual adaptation using non-medical/medical pre-train weights.
Besides, we study the impact of patient ID out-of-distribution on medical visual adaptation, which is a real and challenging scenario. 
Furthermore, results from Med-VTAB indicate that a single pre-trained model falls short in medical task adaptation. 
Therefore, we introduce GMoE-Adapter, a novel method that combines medical and general pre-training weights through a gated mixture-of-experts adapter, achieving state-of-the-art results in medical visual task adaptation.


\end{abstract}

\section{Introduction}

Recent advancements in deep learning have significantly propelled the field of computer vision, especially with the introduction of Vision Transformers (ViTs)~\cite{dosovitskiy2021an,touvron2020deit,liu2021Swin,yuan2021tokens}. 
These models~\cite{he2021masked,caron2021emerging,oquab2023dinov2}, once pre-trained on large-scale datasets, have demonstrated remarkable capabilities across a wide range of visual tasks. 
The adaptability of ViTs to specific downstream tasks through mechanisms such as specialized learnable layers or tokens~\cite{jia2022vpt,yoo2023improving} has opened new avenues for task-specific model optimization. 
This adaptability, known as visual task adaptation, allows for the fine-tuning of pre-trained models to cater to the nuances of particular tasks, enhancing model performance and applicability.

Despite these advancements~\cite{zhai2019VTAB}, the application of visual task adaptation in the medical domain remains underexplored, particularly across diverse medical imaging modalities such as color images, X-rays, and CT scans. 
The medical imaging field presents unique challenges, including but not limited to, the heterogeneity of data, the critical need for high accuracy, and the requirement for models to generalize across different organs and diseases. 
Moreover, the potential of visual task adaptation to leverage pre-existing knowledge from both medical and non-medical domains has not been fully investigated in a systematic and large-scale manner.

To address this gap, we introduce Med-VTAB, a comprehensive Medical Visual Task Adaptation Benchmark designed to facilitate the exploration and evaluation of visual task adaptation techniques in the medical imaging domain. 
Med-VTAB encompasses 1.68 million medical images, spanning a variety of organs and modalities, making it one of the most extensive benchmarks of its kind.
This benchmark is crafted to probe the effectiveness of visual task adaptation strategies and to explore the scaling laws of adaptation in the context of medical imaging. 
Specifically, we examine the relationship between the number of tunable parameters in medical prompt tuning and model performance, alongside assessing the generalizability of adaptations derived from both medical and non-medical pre-training weights.
Furthermore, we investigate the impact of patient ID distribution shifts on the performance of adapted models, a critical aspect in medical applications where model robustness to new patient data is paramount. 

\begin{figure}[t]
\centering
\includegraphics[width=0.75\linewidth]{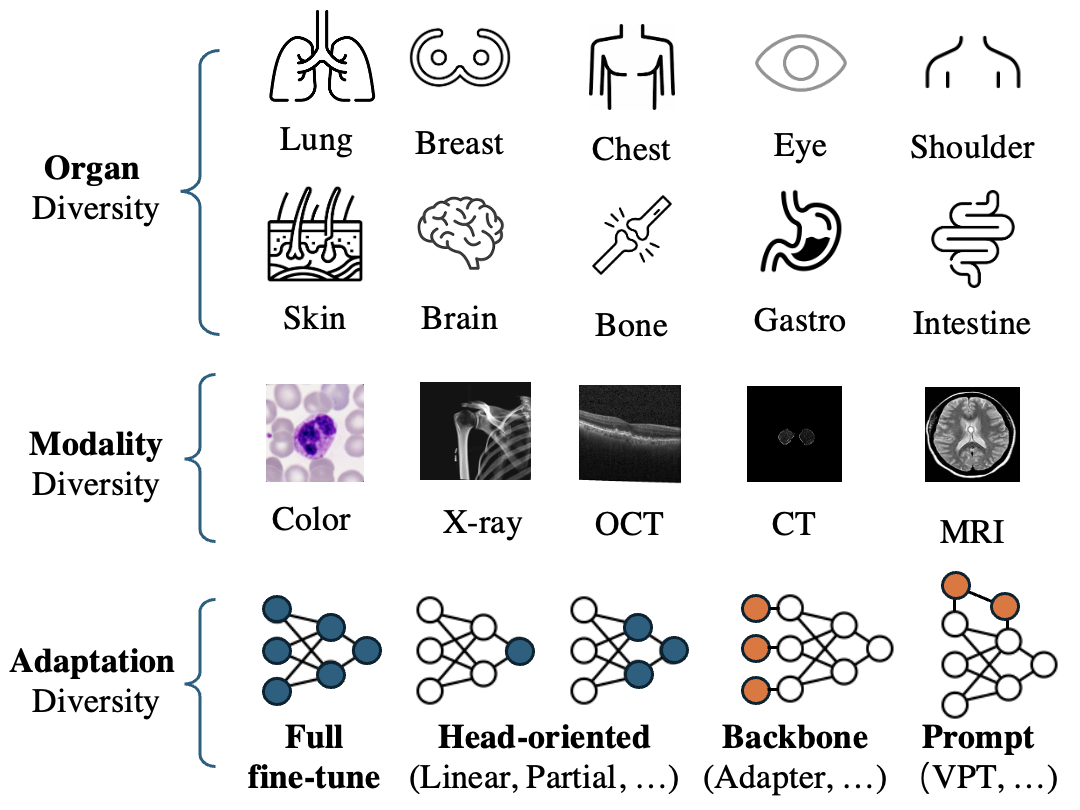}
\caption{Med-VTAB is a large-scale benchmark for adaptation on medical images, consisting of 1.68 million samples, 10 rich organs, and 5 challenging modalities in real-world medical scenarios.
Med-VTAB presents new challenges for impactful adaptation approaches involving full fine-tune, head-oriented (\eg linear probing, partial), backbone-oriented (\eg adapter) and prompt-oriented (\eg VPT) adaptation approaches using pre-trained models from general and medical domains.}
\label{fig: title_img}
\end{figure}

In addition to these explorations, we propose a novel adaptation technique, the Gated Mixture-of-Experts Adapter (GMoE-Adapter), which leverages insights from both medical and general vision pre-training to achieve state-of-the-art performance in medical visual task adaptation. 
The GMoE-Adapter exemplifies the potential of hybrid adaptation strategies that combine domain-specific knowledge with broad, generalist learning from diverse sources.

Through the Med-VTAB benchmark and our investigations into adaptation strategies and scaling laws, this work aims to set a new standard for research in medical visual task adaptation. 
By highlighting the importance of tailored adaptation techniques and the exploration of novel adaptation approaches, we contribute to the advancement of medical imaging analysis, ultimately aiming to enhance diagnostic accuracy and patient outcomes.

Overall, our contributions can be summarized into four main folds:
\begin{itemize}
    \item We present Med-VTAB, a new benchmark encompassing 1.68 million medical images across various organs and modalities, designed to rigorously evaluate the effectiveness of adapting pre-trained ViTs to the domain of medical imaging. 
    \item The introduction of the Gated Mixture-of-Experts Adapter (GMoE-Adapter) marks an improvement in adaptation methodology, combining insights from both medical and general domains to enhance the performance of ViTs on medical visual tasks.
    \item Through meticulous experiments facilitated by Med-VTAB, we provide valuable insights into various adaptation strategies, examining the scaling laws of prompt tuning, the generalizability of model adaptations, and the impact of out-of-distribution scenarios.
    \item Our work culminates in setting new state-of-the-art performance standards on the Med-VTAB benchmark, demonstrating the unparalleled effectiveness and generalizability of the GMoE-Adapter across a broad spectrum of medical imaging modalities and tasks.
\end{itemize}

\section{Related Work}

This section delves into the landscape of existing benchmarks and adaptation methodologies, positioning our work within the broader context of medical task adaptation and highlighting the novel contributions of the Med-VTAB benchmark and the Gated Mixture-of-Experts Adapter (GMoE-Adapter).

\noindent\textbf{Adaptation Benchmark.}
Benchmarks play a crucial role in assessing the performance and generalizability of machine learning models across a wide array of tasks. 
In the domain of visual task adaptation, benchmarks like VTAB (Visual Task Adaptation Benchmark)~\cite{zhai2019VTAB} have set the standard for evaluating the adaptability of models to various visual domains and tasks. However, the specific challenges and requirements of medical imaging—ranging from the diversity of imaging modalities to the critical need for high accuracy in diagnostic tasks—necessitate a dedicated benchmark. 
Our Med-VTAB fills this gap by providing a comprehensive suite of medical imaging datasets that encompass a wide range of organs, modalities, and diseases, specifically tailored for the evaluation of visual task adaptation techniques in the medical domain. 
Unlike existing benchmarks that predominantly focus on general visual tasks, Med-VTAB's emphasis on medical images offers a targeted platform for advancing medical image analysis through adapted pre-trained models.

\noindent\textbf{Medical Task Adaptation.}
Adaptation techniques in medical imaging have evolved from simple fine-tuning approaches to more sophisticated methods designed to leverage pre-trained models without extensive retraining. 
Previous works~\cite{rasmy2020medbert,wang2022medclip,xiao2023delving,yang2023mrm,nguyen2023lvm} in medical task adaptation primarily focused on transfer learning, where models pre-trained on large medical datasets were fine-tuned for specific medical tasks on small datasets. 
While effective, this approach often required substantial computational resources and large annotated medical datasets, which are not always available.
Recent advances have introduced more nuanced adaptation strategies, such as adapters~\cite{pfeiffer2020adapterfusion}, where only smaller layers of a pre-trained model are trained on medical data, and prompt tuning~\cite{jia2022vpt,yoo2023improving}, which adjusts model inputs to guide the adaptation process. 
However, these methods still face challenges in fully exploiting the rich representations learned by models pre-trained on diverse visual tasks.
The GMoE-Adapter represents a significant leap forward in medical task adaptation. By integrating a gated mixture-of-experts mechanism within the adaptation process, it allows for dynamic, context-aware adjustment of model features to better suit specific medical tasks. 
This approach not only enhances model performance across a variety of medical imaging tasks but also introduces a novel framework for understanding and optimizing the adaptation process in the context of medical imaging.

\section{Med-VTAB: Medical Visual Task Adaptation Benchmark}

To comprehensively evaluate the efficacy of visual task adaptation in the medical domain, we developed the Medical Visual Task Adaptation Benchmark (Med-VTAB). 
This benchmark is meticulously designed to cover a broad spectrum of medical images (1.68 million), encompassing diverse organs and modalities, and to test various adaptation approaches. 
Below, we detail the benchmark's structure in terms of organ diversity, modality diversity, and the range of adaptation methods explored.

\subsection{Organ Diversity}

\begin{figure}[t]
\centering
\includegraphics[width=0.95\linewidth]{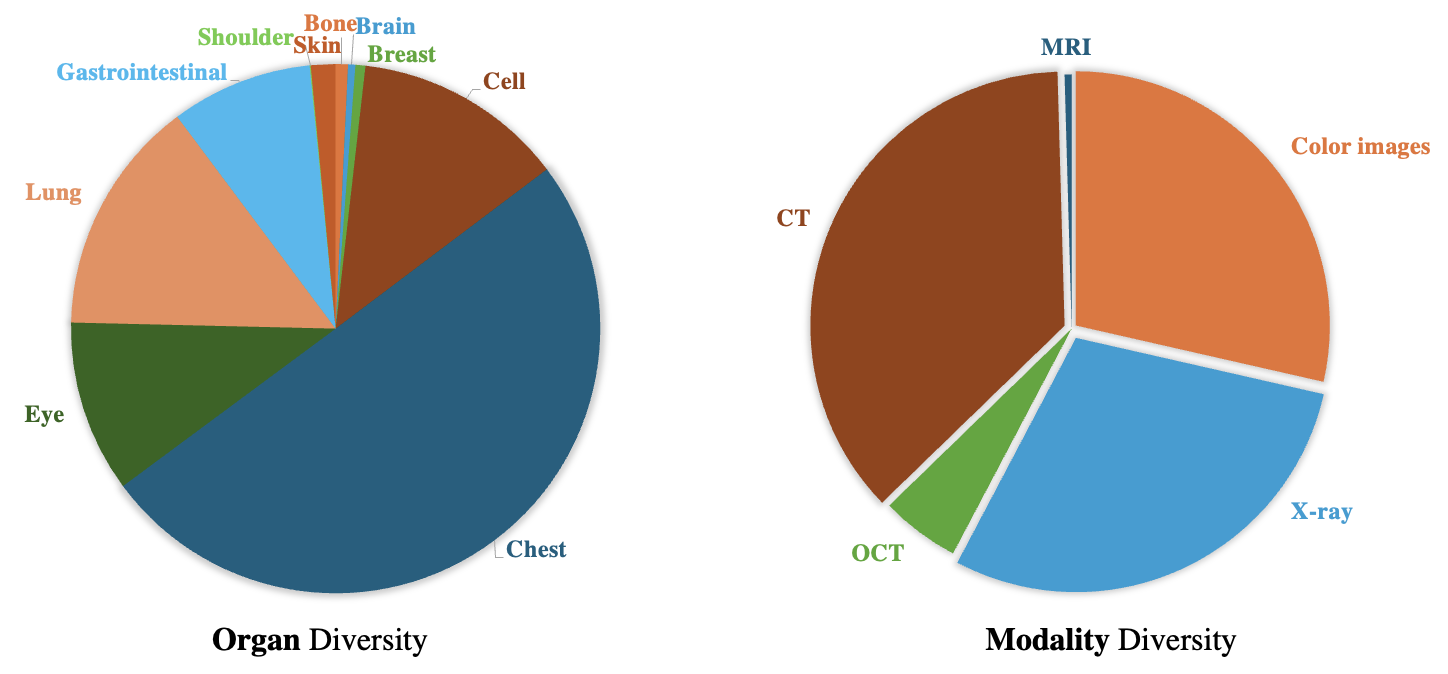}
\caption{Statistics of organ diversity (Left, 10 rich organs) and modality diversity (Right, 5 challenging modalities) in the proposed Medical Visual Task Adaptation Benchmark (Med-VTAB) consisting of 1.68 million images.}
\label{fig: stat_img}
\end{figure}

Med-VTAB includes a wide array of organ-specific datasets, ensuring comprehensive coverage of the human body and a variety of diseases and conditions. 
The inclusion of diverse organs enables the evaluation of adaptation techniques across different anatomical structures and pathologies. The organs covered in the benchmark are as follows: Lung, Breast, Chest, Eye, Shoulder, Skin, Brain, Bone, and Gastrointestinal, as shown in Fig.~\ref{fig: stat_img} (Left).
This diversity ensures that the benchmark tests not only the generalizability of the adaptation methods but also their applicability to specific medical scenarios.

To accommodate the fixed input size of Vision Transformers, images from all datasets are resized. However, care is taken to preserve the aspect ratio and diagnostic integrity of the images, crucial in medical imaging.
Intensity normalization is applied to account for variations in lighting, contrast, and exposure across different imaging modalities and equipment. 
This step ensures that the models focus on relevant anatomical or pathological features rather than imaging artifacts.
To enhance the robustness of the models and mitigate overfitting, data augmentation techniques such as rotations, flips, and slight zooms are employed. These augmentations are carefully selected to be medically plausible, avoiding transformations that could introduce misleading or non-physiological features.
We will provide more details in the supplementary material.

\subsection{Modality Diversity}

Medical imaging employs a variety of modalities, each offering unique insights into the body's internal structures. 
Med-VTAB reflects this diversity by including datasets from the following imaging modalities: Color Images, X-ray, Optical Coherence Tomography (OCT), Computed Tomography (CT), and Magnetic Resonance Imaging (MRI), as shown in Fig.~\ref{fig: stat_img} (Right).
This range allows for the assessment of adaptation methods across different imaging techniques, highlighting their versatility and performance in handling the specific challenges posed by each modality, such as varying image resolutions, contrasts, and dimensions.
Certain modalities, such as MRI and CT, benefit from additional preprocessing steps like windowing, which adjusts the range of pixel intensities to highlight specific tissues or structures. These modality-specific adjustments are critical for maximizing the diagnostic value of the images.

The preprocessing steps adopted in Med-VTAB are designed to be both comprehensive and reasonable, ensuring that the models are trained and evaluated on data that closely resemble real-world clinical imaging scenarios. The comprehensiveness of the preprocessing pipeline ensures that a wide array of imaging modalities and diagnostic tasks are adequately represented, providing a robust foundation for assessing the generalizability and adaptability of visual task adaptation techniques.
Furthermore, the reasonableness of the preprocessing steps is evident in their alignment with clinical imaging practices and their focus on preserving the diagnostic integrity of the images. By incorporating modality-specific processing steps and employing augmentation techniques judiciously, the preprocessing pipeline respects the clinical context of the images, ensuring that the resulting models are both accurate and clinically relevant.

\subsection{Adaptation Diversity}

A core aspect of Med-VTAB is its exploration of various adaptation strategies. 
These strategies are categorized into full fine-tune, head-oriented, backbone-oriented, and prompt-oriented adaptations, each targeting different components of the pre-trained models for optimization. 
The specific approaches explored under each category are as follows:

\noindent\textbf{Full fine-tune.}
This approach involves the comprehensive fine-tuning of all model parameters, serving as a baseline for adaptation performance.

\noindent\textbf{Head-oriented.}
This category focuses on adapting the final layers (heads) of the model, with strategies including:
1) Linear prob: finetune a single linear layer as a classification head;
2) MLP-3: use a three-layer Multi-Layer Perceptron adaptation as the classification head;
3) Partial-1~\cite{noroozi2016unsupervised}: only fine-tune the last block of the backbone and freeze other blocks.

\noindent\textbf{Backbone-oriented.}
Targeting the model's backbone, these methods aim to adjust the core representations:
1) Sidetune~\cite{zhang2020sidetuning}: introducing parallel networks that are fine-tuned while the main network remains frozen;
2) Bias~\cite{ben2022bitfit}: adjusting only the bias terms in the model.
3) Adapter~\cite{pfeiffer2020adapterfusion}: inserting small, trainable layers within the model's architecture.

\noindent\textbf{Prompt-oriented.}
Leveraging the concept of prompts, these methods introduce small modifications or additional inputs to guide the model's focus:
1) VPT~\cite{jia2022vpt}-Shallow \& -Deep: Variants of Visual Prompt Tuning that introduce prompts at different depths of the model;
2) GaPT~\cite{yoo2023improving}: a gated prompt tuning approach that dynamically adjusts prompts based on the input.
3) LSPT~\cite{mo2024lspt}: a strong prompt tuning method with long-term gated prompts on both temporal and spatial coding.

By encompassing this broad range of adaptation strategies, Med-VTAB enables a thorough investigation into the scalability, efficiency, and effectiveness of various adaptation methods in the context of medical imaging. 
Through this benchmark, we aim to identify optimal strategies for leveraging pre-trained models across diverse medical visual tasks, thereby advancing the field of medical image analysis and improving diagnostic capabilities.

\section{Method}

The divergence between the medical and general domains presents a significant challenge for models trained exclusively within one domain, as those focused solely on general domain lack the specificity required for medical applications, whereas models dedicated to the medical domain suffer from limitations in training data volume and model capacity. 
To bridge this gap, we consider leveraging multiple Vision Transformers (ViTs) for adapation, and introduce the Gated Mixture-of-Experts Adapter (GMoE-Adapter), a novel approach aimed at enhancing the adaptation of ViTs for a broad spectrum of medical imaging tasks.
In this section, we first outline the notations used and revisit the concept of adapters in ViTs. Then, we present methods on leveraging multiple ViTs, including an intuitive baseline and the proposed GMoE-Adapter.

\subsubsection{Notations}
Let $X$ denote the input image, and $Y$ the corresponding label in a medical imaging dataset. 
We use $f(\cdot)$ to represent the pre-trained ViT model, which transforms $X$ into a set of feature representations. The adapter module, denoted as $A(\cdot)$, is applied to these representations to produce task-specific features. 
The final output, $\hat{Y}$, represents the model's prediction. Parameters of $f(\cdot)$ are denoted by $\Theta$, while $\Phi$ represents the parameters of the adapter module.

\subsubsection{Adapter}
Adapters~\cite{pfeiffer2020adapterfusion} are small neural network modules inserted between the layers of a pre-trained model. 
They allow for fine-tuning on specific tasks without altering the original model parameters significantly. 
This approach is particularly appealing for medical imaging, where the amount of available training data may be limited, and preserving the knowledge in pre-trained models is crucial. 
Adapters typically consist of a few fully connected layers and non-linear activation functions, enabling the model to learn task-specific features efficiently.

\subsection{Leveraging Multiple Expert Pre-Trained Models}

\subsubsection{Mixture-of-Experts Adapter}

The Mixture-of-Experts (MoE) Adapter extends the adapter concept by incorporating multiple expert networks within the adapter module. 
Each expert for general $A_g(\cdot)$ and medical adapter $A_m(\cdot)$ is designed to specialize in different aspects or features of the input data. 
The MoE approach allows for a more nuanced adaptation of the model to the task at hand by combining the outputs of these experts based on the input data, that is, $\hat{Y}_{moe} = A_g(\cdot) + A_m(\cdot)$.

\subsubsection{Gated Mixture-of-Experts Adapter}

\begin{figure}[t]
\centering
\includegraphics[width=0.95\linewidth]{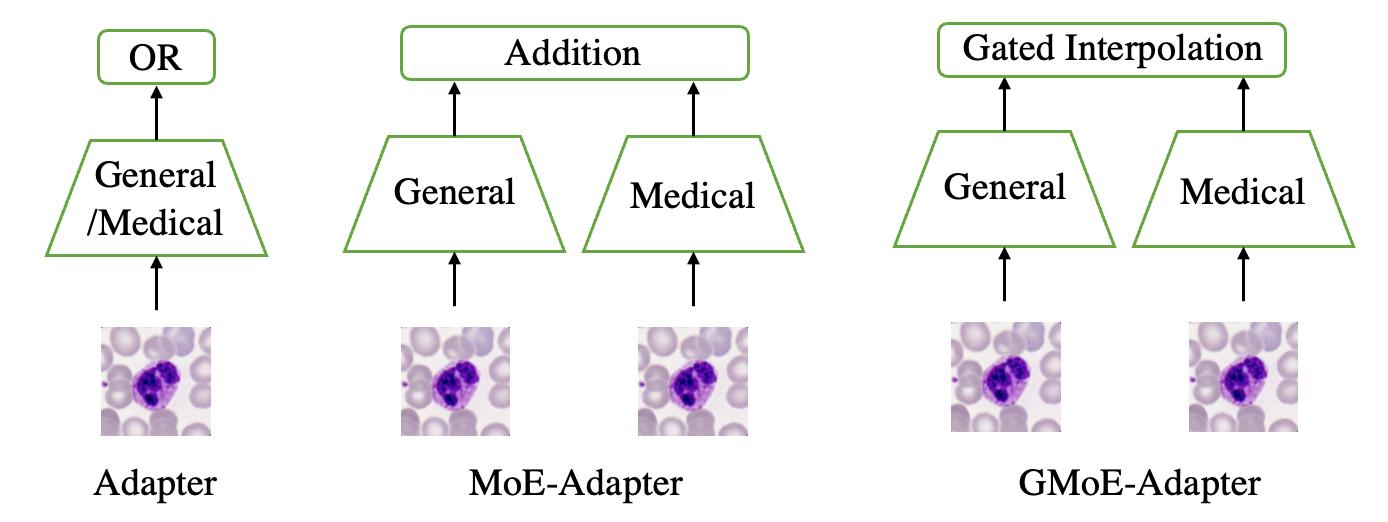}
\caption{Illustration of the proposed Gated Mixture-of-Experts (GMoE)-Adapter framework versus the standard adapter and MoE-adapter methods.}
\label{fig: main_img}
\end{figure}

Building upon the MoE Adapter, the Gated Mixture-of-Experts Adapter (GMoE-Adapter) introduces a gating mechanism to control the contribution of each expert to the final adapted feature representation, as shown in Fig.~\ref{fig: main_img}.
The gating mechanism is a trainable component that determines the relevance of each expert for a given input, enabling the model to dynamically allocate computational resources and attention to the most pertinent experts.

The GMoE-Adapter comprises two key components:
1) Experts: pre-trained backbones from two different domains, each designed to capture different aspects of the medical imaging data using general $A_g(\cdot)$ and medical $A_m(\cdot)$ adapters.
2) Gating: a learnable parameter $\alpha$ that computes weights for each expert based on the input features, effectively determining which experts are most relevant for the current task.
The final output for GMoE-Adapter is defined with respect to the gated interpolation of each expert adapter output as $\hat{Y}_{\text{GMoE}} = \alpha \cdot A_g(\cdot) + (1-\alpha) \cdot A_m(\cdot)$,
where $\alpha\in\mathbb{R}^{1\times D}$ and $D$ is the dimension of embeddings.

The integration of the GMoE-Adapter into pre-trained ViTs involves inserting the adapter module at strategic points within the model, typically after each transformer block. 
This placement allows the adapted model to refine the pre-trained features iteratively, ensuring that the final feature representations are highly tailored to the specific requirements of medical imaging tasks.
Through the GMoE-Adapter, we aim to harness the rich representational power of pre-trained ViTs and enhance their adaptability to the nuanced and diverse challenges of medical image analysis.

\section{Benchmarking Medical Visual Task Adaptation}

\subsection{Experimental Setup}

We detail our experimental setup, including the datasets used, evaluation metrics, implementation details, benchmarking results on medical visual task adaptation, and comparisons with prior work.

\paragraph{Datasets.} 

Our experiments span a comprehensive set of medical image datasets to ensure a broad evaluation of our GMoE-Adapter across various organs and modalities.
Specifically, we utilize nine color medical image datasets that encompass a wide range of medical imaging challenges, including polyp (HyperKvasir~\cite{borgli2020HyperKvasir}, MESAD Prostatectomy~\cite{Bawa2021ESAD}, Kvasir~\cite{Kvasirv2}), cell (AMLC~\cite{matek2019human}, LHNCBC~\cite{Lhncbc}, MLLBone~\cite{Matek2021highly}), eye (APTOS~\cite{aptos}, EyePACS~\cite{eyepacs}), and skin (ISIC~\cite{isic}).
We also collected seven X-ray image datasets from diverse organs covering lung (Vindr~\cite{Nguyen2022vindr}, COVIDx~\cite{wang2020covidx}, RSNA~\cite{shih2019rsna}), 
breast (CBIS~\cite{lee2017cbis}), shoulder (SYMH~\cite{symh}), bone (RSNA Bone~\cite{Halabi2019TheRP}), and chest (CheXpert~\cite{Irvin2019chexpert}).
We included OCT, CT, and MRI modalities in seven other datasets including eye (Heidelberg~\cite{Kermany2018LargeDO}), chest (CC-CCII~\cite{zhang2020clinically}, Mosmed~\cite{Morozov2020MosMedDataCC}, COVID-C~\cite{RAHIMZADEH2021102588}, RICORD~\cite{tsai2021ricord}) and brain (PPMI~\cite{Kenneth2011ppmi}, Brain-Tumor~\cite{braintumor}).
These datasets represent a diverse array of medical imaging tasks, from macroscopic organ imaging to microscopic cell analysis, enabling us to test existing adaptation methods and our GMoE-Adapter's adaptability and performance comprehensively.

\paragraph{Evaluation Metrics.}
We employ the average accuracy score as the primary metric for evaluating model performance on each dataset. 
When evaluated on X-ray images, the performance of Area under ROC curve (AUROC) are reported.
The accuracy is calculated over the test set, and to ensure robustness of the results, we report the average score across three runs, each initialized with different random seeds.

\paragraph{Implementation.}
We implement our GMoE-Adapter within the architecture of pre-trained Vision Transformers (ViTs), specifically using DINO v2~\cite{oquab2023dinov2} pre-trained ViT-B/16 models (on 1.28 million general images) and medical pre-trained models~\cite{nguyen2023lvm} (on 1.6 million cell images) as our base. 
We use the same pre-trained model parameters as previous work~\cite{jia2022vpt}.

\begin{table*}[t]
	\renewcommand\tabcolsep{6.0pt}
	\centering
        \caption{Quantitative results of visual prompt tuning of SSL pre-trained vision transformers (MAE~\cite{he2021masked}, MoCo v3~\cite{chen2021mocov3}, DINO v2~\cite{oquab2023dinov2}) on color images. Total Params denotes the total number of parameters for the backbone encoder ViT-B, prompt tokens, and the task heads.}
   \label{tab: exp_sota_color}
	\scalebox{0.6}{
		\begin{tabular}{lcccccccccc}
			\toprule
			\multirow{2}{*}{Method}     & Total & HyperKvasir & MESAD & AMLC & APTOS & ISIC & Kvasir & LHNCBC & MLLBone & EyePACS \\
            & Params & Polyp & Prostatectomy & Cell & Eye & Skin & Polyp & Cell & Cell & Eye \\
			\midrule
                \multicolumn{10}{l}{\it MAE Pre-train ViT-B/16:} \\
                Full        & 19.01X & \bf 73.25               & \bf 52.39                 & \bf 45.23        & \bf 68.59       & \bf 58.67       & \bf 87.65          & \bf 51.36         & \bf 47.36          & \bf 63.75         \\ 
                Linear      & 1.01X  & 43.26               & 21.75                 & 18.52        & 37.56       & 35.23       & 52.16          & 26.53         & 20.17          & 35.19         \\
                Adapter~\cite{pfeiffer2020adapterfusion}     & 1.17X  & \underline{67.35}               & \underline{45.62}                 & \underline{38.18}        & \underline{60.17}       & \underline{52.15}       & \underline{80.56}          & \underline{46.18}         & \underline{40.76}          & \underline{58.63}         \\ 
                VPT-Shallow~\cite{jia2022vpt} & 1.02X  & 52.32               & 33.16                 & 27.29        & 49.76       & 42.53       & 59.06          & 34.25         & 23.29          & 42.16         \\
                VPT-Deep~\cite{jia2022vpt}    & 1.02X  & 56.13               & 37.53                 & 30.51        & 53.62       & 46.79       & 63.17          & 39.32         & 29.86          & 45.21         \\
                GaPT~\cite{yoo2023improving}        & 1.02X  & 59.32               & 41.67                 & 31.68        & 56.13       & 48.53       & 65.25          & 41.53         & 31.96          & 49.32         \\
                LSPT~\cite{mo2024lspt}        & 1.05X  & 61.67               & 43.26                 & 33.25        & 58.92       & 50.17       & 68.17          & 43.78         & 33.25          & 51.28        \\ \hline
                \multicolumn{10}{l}{\it MoCo v3 Pre-train ViT-B/16:} \\
                FULL        & 19.01X & \bf 75.37               & \bf 54.72                 & \bf 46.78        & \bf 69.32       & \bf 60.15       & \bf 88.25          & \bf 53.76         & \bf 48.82          & \bf 65.62         \\
                Linear      & 1.01X  & 47.75               & 28.65                 & 22.53        & 42.76       & 39.86       & 56.31          & 29.85         & 25.73          & 40.16         \\
                Adapter~\cite{pfeiffer2020adapterfusion}     & 1.17X  & \underline{68.52}               & \underline{47.25}                 & \underline{39.85}        & \underline{62.13}       & \underline{53.68}       & \underline{82.28}          & \underline{47.53}         & \underline{42.15}          & \underline{59.19}         \\
                VPT-Shallow~\cite{jia2022vpt} & 1.02X  & 55.82               & 35.78                 & 29.15        & 51.67       & 45.06       & 61.23          & 36.82         & 26.57          & 44.26         \\
                VPT-Deep~\cite{jia2022vpt}    & 1.02X  & 58.27               & 39.05                 & 32.81        & 54.78       & 48.65       & 65.21          & 40.05         & 31.02          & 46.79         \\
                GaPT~\cite{yoo2023improving}         & 1.02X  & 61.24               & 42.31                 & 34.57        & 57.09       & 49.72       & 66.02          & 42.36         & 33.52          & 50.23         \\
                LSPT~\cite{mo2024lspt}        & 1.05X  & 63.57               & 44.52                 & 35.72        & 59.33       & 51.26       & 68.73          & 45.21         & 35.71          & 51.76        \\ \hline
                \multicolumn{10}{l}{\it DINO v2 Pre-train ViT-B/16:} \\
                FULL        & 19.01X & \bf 76.53               & \bf 56.17                 & \bf 48.39        & \bf 70.26       & \bf 61.25       & \bf 89.32          & \bf 56.87         & \bf 50.21          & \bf 67.23         \\
                Linear      & 1.01X  & 51.67               & 32.16                 & 25.63        & 45.72       & 42.36       & 58.85          & 32.17         & 28.65          & 42.37         \\
                Adapter~\cite{pfeiffer2020adapterfusion}     & 1.17X  & \underline{70.38}               & \underline{49.75}                 & \underline{42.16}        & \underline{65.38}       & \underline{55.19}       & \underline{83.57}          & \underline{49.78}         & \underline{43.86}          & \underline{60.82}         \\
                VPT-Shallow~\cite{jia2022vpt} & 1.02X  & 59.76               & 39.75                 & 31.62        & 53.95       & 47.32       & 63.72          & 38.53         & 30.26          & 46.58         \\
                VPT-Deep~\cite{jia2022vpt}    & 1.02X  & 62.89               & 43.78                 & 35.75        & 57.52       & 50.89       & 66.53          & 42.87         & 35.37          & 48.75         \\
                GaPT~\cite{yoo2023improving}        & 1.02X  & 65.18               & 45.79                 & 37.26        & 59.37       & 51.58       & 67.13          & 45.16         & 36.85          & 51.57         \\
                LSPT~\cite{mo2024lspt}        & 1.05X  & 67.23               & 47.53                 & 38.72        & 61.25       & 53.62       & 69.79          & 47.51         & 38.92          & 52.86     \\   
			\bottomrule
			\end{tabular}}
\end{table*}

\begin{table*}[t]
	\renewcommand\tabcolsep{6.0pt}
	\centering
        \caption{Quantitative results of visual prompt tuning of DINO v2 pre-trained vision transformers on X-ray images. Total Params denotes the total number of parameters for the backbone encoder ViT-B, prompt tokens, and the task heads.}
   \label{tab: exp_sota_xray}
	\scalebox{0.7}{
		\begin{tabular}{lcccccccc}
			\toprule
			\multirow{2}{*}{Method}     & Total & Vindr & CBIS   & COVIDx & SYMH     & RSNA Bone & CheXpert & RSNA \\
            & Params & Lung  & Breast & Lung   & Shoulder & Bone      & Chest    & Lung \\
			\midrule
                FULL        & 19.01X &  \bf 75.22 & \bf 86.51 & \bf 87.93 & \bf 85.36 & \bf 62.62 & \bf 82.75 & \bf 78.16 \\
                Linear      & 1.01X  &  62.81 & 71.32 & 72.56 & 72.81 & 46.73 & 67.26 & 65.38 \\
                Adapter~\cite{pfeiffer2020adapterfusion}     & 1.17X  &  \underline{70.35} & \underline{81.26} & \underline{80.72} & \underline{79.52} & \underline{55.35} & \underline{73.61} & \underline{72.93} \\
                VPT-Shallow~\cite{jia2022vpt} & 1.02X  & 63.56 & 72.23 & 73.83 & 74.35 & 50.21 & 69.73 & 67.69 \\
                VPT-Deep~\cite{jia2022vpt}    & 1.02X  & 65.73 & 74.61 & 76.18 & 76.86 & 51.72 & 70.85 & 69.25 \\
                GaPT~\cite{yoo2023improving}         & 1.02X  &  66.92 & 75.15 & 77.25 & 77.25 & 52.83 & 71.37 & 70.29 \\
                LSPT~\cite{mo2024lspt}        & 1.05X  & 67.87 & 76.23 & 78.33 & 77.96 & 53.51 & 71.92 & 70.86 \\
			\bottomrule
			\end{tabular}}
\end{table*}

\begin{table*}[t]
	\renewcommand\tabcolsep{6.0pt}
	\centering
        \caption{Quantitative results of visual prompt tuning of DINO v2 pre-trained vision transformers on OCT, CT, and MRI images. Total Params denotes the total number of parameters for the backbone encoder ViT-B, prompt tokens, and the task heads.}
   \label{tab: exp_sota_other}
	\scalebox{0.7}{
		\begin{tabular}{lcccccccc}
			\toprule
			\multirow{2}{*}{Method}     & Total & Heidelberg & CC-CCII & Mosmed & COVID-C & RICORD & PPMI  & Brain-Tumor \\
            & Params & Eye        & Chest   & Chest  & Chest   & Chest  & Brain & Brain  \\
			\midrule
                FULL        & 19.01X & \bf 72.16 & \bf 74.82 & \bf 76.83 & \bf 77.52 & \bf 75.82 & \bf 72.63 & \bf 74.53 \\
                Linear      & 1.01X  & 63.25 & 60.87 & 62.87 & 60.93 & 58.35 & 55.27 & 62.35 \\
                Adapter~\cite{pfeiffer2020adapterfusion}     & 1.17X  & \underline{67.58} & \underline{66.23} & \underline{65.52} & \underline{66.37} & \underline{64.21} & \underline{61.35} & \underline{67.62} \\
                VPT-Shallow~\cite{jia2022vpt} & 1.02X  & 64.15 & 60.75 & 63.21 & 61.05 & 59.07 & 56.35 & 62.75 \\
                VPT-Deep~\cite{jia2022vpt}    & 1.02X  & 64.78 & 61.26 & 63.65 & 61.78 & 59.53 & 56.93 & 63.37 \\
                GaPT~\cite{yoo2023improving}         & 1.02X  & 65.06 & 61.37 & 63.69 & 61.95 & 59.71 & 56.97 & 63.52 \\
                LSPT~\cite{mo2024lspt}        & 1.05X  & 65.23 & 61.56 & 63.75 & 62.12 & 59.85 & 57.08 & 63.67 \\
			\bottomrule
			\end{tabular}}
\end{table*}

\subsection{Main Results}\label{sec:exp_sota}

To comprehensively assess the capabilities of existing adaptation methods in medical visual task adaptation, we performed extensive benchmarking across various medical imaging modalities. 
This section reports on the performance of existing visual task adaptation methods across color images, X-ray, OCT, CT, and MRI modalities.

For color image datasets, including those for skin lesion analysis and retinal disease diagnosis, our GMoE-Adapter demonstrated superior performance in adapting pre-trained ViTs to these specific tasks. The results, detailed in Table~\ref{tab: exp_sota_color}, highlight the adapter's effectiveness in capturing the nuanced features required for accurate diagnosis from color medical images by inserting small and trainable parameters into the backbone.

The X-ray modality, crucial for diagnosing a wide range of conditions from bone fractures to lung diseases, posed unique challenges due to its high variance in image density and structure. 
As shown in Table~\ref{tab: exp_sota_xray}, the Adapter still outperforms linear probing and previous prompt tuning approaches with limited trainable parameters, showcasing its ability to enhance model sensitivity to relevant pathological features in X-ray images.
For other modalities such as OCT, CT, and MRI, which are characterized by their depth and complexity of the captured images, the Adapter again proved its mettle. 
Table~\ref{tab: exp_sota_other} encapsulates the adapter's performance, demonstrating its versatility and effectiveness across these technologically diverse imaging techniques.

\begin{table*}[t]
	\renewcommand\tabcolsep{6.0pt}
	\centering
        \caption{Quantitative results of parameter-efficient adapters from DINO v2 and Medical domain pre-trained vision transformers. Total Params denotes the total number of parameters for the backbone encoder ViT-B, prompt tokens, and the task heads.}
   \label{tab: exp_sota_cmp}
	\scalebox{0.57}{
		\begin{tabular}{lcccccccccc}
			\toprule
			\multirow{2}{*}{Method}     & Total & HyperKvasir & MESAD & AMLC & APTOS & ISIC & Kvasir & LHNCBC & MLLBone & EyePACS \\
            & Params & Polyp & Prostatectomy & Cell & Eye & Skin & Polyp & Cell & Cell & Eye \\
			\midrule
                Linear    & 1.01X  & 51.67 & 32.16 & 25.63 & 45.72 & 42.36 & 58.85 & 32.17 & 28.65 & 42.37 \\
                MLP-3     & 1.35X  & 53.79 & 33.17 & 28.95 & 47.21 & 45.25 & 60.53 & 35.67 & 30.78 & 43.98 \\
                Partial-1~\cite{noroozi2016unsupervised} & 3.00X  & 57.21 & 35.72 & 32.12 & 49.52 & 47.18 & 63.18 & 39.15 & 32.85 & 45.26 \\
                Sidetune~\cite{zhang2020sidetuning}  & 3.69X  & 49.23 & 31.25 & 24.86 & 43.15 & 40.21 & 55.12 & 30.73 & 26.29 & 40.65 \\
                Bias~\cite{ben2022bitfit}      & 1.05X  & 68.53 & 45.12 & 40.15 & 62.19 & 53.12 & 79.58 & 47.92 & 41.85 & 58.16 \\
                Adapter~\cite{pfeiffer2020adapterfusion} (DINO v2)   & 1.17X  & 70.38 & 49.75 & 42.16 & 65.38 & 55.19 & 83.57 & 49.78 & 43.86 & 60.82 \\
                Adapter~\cite{pfeiffer2020adapterfusion} (Medical)   & 1.17X  & 70.29 & 49.72 & 42.37 & 65.23 & 55.02 & 83.35 & 50.26 & 44.25 & 60.78 \\
                MoE-Adapter   & 1.34X  & 70.42 & 49.78 & 42.47 & 65.39 & 55.21 & 83.58 & 50.35 & 44.32 & 60.89 \\
                GMoE-Adapter   & 1.35X  & \bf 70.75 & \bf 50.26 & \bf 42.83 & \bf 65.51 & \bf 55.37 & \bf 83.79 & \bf 50.86 & \bf 44.75 & \bf 61.02 \\
                                
			\bottomrule
			\end{tabular}}
\end{table*}

\subsection{Evaluation of GMoE-Adapter}\label{sec:exp_sota}

In comparing our GMoE-Adapter to prior works, including the original Mixture-of-Experts Adapter and other state-of-the-art adaptation techniques, we further establish the superiority of our approach. The comparative analysis, summarized in Table~\ref{tab: exp_sota_cmp}, illustrates the enhanced performance of the GMoE-Adapter across all evaluated datasets and modalities.
In particular, the proposed GMoE-Adapter significantly outperforms the vanilla Adapter based on DINO v2~\cite{oquab2023dinov2} by 0.37, 0.51, 0.67 on Polyp, Prostatectomy, and Cell images.
Moreover, we achieve superior performance gains compared to head-oriented baselines, such as linear probing and Sidetune~\cite{zhang2020sidetuning}.
Meanwhile, our GMoE-Adapter outperforms MLP-3 by a large margin although both have comparable trainable parameters.
These significant improvements demonstrate the superiority of our method in medical image classification adaptation. 

In addition, we achieve significant gains compared to the vanilla Adapter based on medical backbone from 1.6 million cell images pre-training.
Compared to MoE-Adapter, we also achieve highly better results regarding all medical color images.
While the MoE-Adapter laid the groundwork for incorporating expert knowledge into the adaptation process, our GMoE-Adapter builds upon this foundation by introducing a gating mechanism. 
This innovation allows for dynamic selection and combination of expert selection based on the specific features of each input image, leading to more precise and accurate model predictions.

\section{In-Depth Analysis for Medical Visual Task Adaptation}

In this section, we introduce the experiments conducted by us to discuss the following research insights:

\paragraph{Insights 1.} Scaling law of medical prompt tuning concerning tunable parameters.

\paragraph{Insights 2.} Generalizability of medical visual adaptation using non-medical and medical pre-train weights.

\paragraph{Insights 3.} Patient ID Out-of-Distribution of medical visual adaptation.

\begin{table*}[t]
	\renewcommand\tabcolsep{6.0pt}
	\centering
        \caption{Scaling law of medical visual prompt tuning of DINO v2 pre-trained vision transformers on color images. Total Params denotes the total number of parameters for the backbone encoder ViT-B, prompt tokens, and the task heads.}
   \label{tab: exp_scaling}
	\scalebox{0.65}{
		\begin{tabular}{cccccccccc}
			\toprule
			Total & HyperKvasir & MESAD & AMLC & APTOS & ISIC & Kvasir & LHNCBC & MLLBone & EyePACS \\
            Params & Polyp & Prostatectomy & Cell & Eye & Skin & Polyp & Cell & Cell & Eye \\
			\midrule
                1.39X & \bf 64.97 & \bf 45.37 & \bf 37.29 & \bf 59.35 & \bf 52.37 & \bf 68.05 & \bf 44.87 & \bf 37.93 & \bf 51.86 \\
                1.17X & 64.15 & 44.52 & 36.56 & 58.63 & 51.52 & 67.63 & 43.95 & 37.21 & 50.73 \\
                1.10X & 63.96 & 44.17 & 36.23 & 58.25 & 51.06 & 67.21 & 43.59 & 36.62 & 50.12 \\
                1.05X & 63.25 & 43.95 & 35.97 & 57.83 & 50.96 & 66.92 & 43.21 & 35.81 & 49.29 \\
                1.02X & 62.89 & 43.78 & 35.75 & 57.52 & 50.89 & 66.53 & 42.87 & 35.37 & 48.75 \\
                1.01X & 62.21 & 43.17 & 35.23 & 57.18 & 50.22 & 66.12 & 42.35 & 34.87 & 48.15 \\
			\bottomrule
			\end{tabular}}
\end{table*}

\subsection{Scaling Law of Medical Prompt Tuning}

Inspired by the Scaling Law study~\cite{Gao2022ScalingLF} by OpenAI, we investigate the scaling law of medical prompt tuning in the context of tunable parameters. Our analysis involves adjusting the number of tunable parameters and observing the corresponding performance changes on various medical imaging tasks. 
The results, detailed in Table~\ref{tab: exp_scaling}, illustrate a clear trend (\textit{e.g.}, 62.21$\rightarrow$ 64.97 on Polyp) where increasing the number of tunable parameters from 1.01X to 1,39X enhances the model's performance. 
When it comes to other organs, a similar increasing trend of performance can be observed with the increase in the number of tunable parameters.
This finding suggests a desirable scaling law between the number of prompt tuning parameters and the effectiveness of adaptation, highlighting the efficiency of prompt tuning for medical applications.

\begin{table*}[t]
	\renewcommand\tabcolsep{6.0pt}
	\centering
        \caption{Generalizability results of visual prompt tuning of medical/non-medical pre-trained vision transformers on color images. Total Params denote the total number of parameters for the backbone encoder ViT-B, prompt tokens, and the task heads.}
   \label{tab: exp_general}
	\scalebox{0.6}{
		\begin{tabular}{lcccccccccc}
			\toprule
			\multirow{2}{*}{Pre-train Data}     & Total & HyperKvasir & MESAD & AMLC & APTOS & ISIC & Kvasir & LHNCBC & MLLBone & EyePACS \\
            & Params & Polyp & Prostatectomy & Cell & Eye & Skin & Polyp & Cell & Cell & Eye \\
			\midrule     
            Medical     & 1.02X & 62.57 & 43.26 & 36.21 & 57.36 & 50.75 & 66.34 & 43.28 & 35.82 & 48.68 \\
            Non-medical & 1.02X & 62.89 & 43.78 & 35.75 & 57.52 & 50.89 & 66.53 & 42.87 & 35.37 & 48.75 \\
            Medical     & 1.17X & 64.12 & 44.39 & \bf 36.58 & 58.52 & 51.39 & 67.32 & \bf 44.06 & \bf 37.36 & 50.26 \\
            Non-medical & 1.17X & \bf 64.15 & \bf 44.52 & 36.56 & \bf 58.63 & \bf 51.52 & \bf 67.63 & 43.95 & 37.21 & \bf 50.73 \\
			\bottomrule
			\end{tabular}}
\end{table*}

\subsection{Generalizability of Medical Visual Adaptation}

The generalizability of medical visual adaptation is examined by applying visual prompt tuning~\cite{jia2022vpt} to models pre-trained on both medical and non-medical datasets.
Specifically, we use DINO v2~\cite{oquab2023dinov2} pre-trained ViT-B/16 models on 1.28 million general images and medical pre-trained models~\cite{nguyen2023lvm} on 1.6 million cell images.
This experiment conducted on all color image modalities aims to understand how the source of pre-training influences the adaptation process and the resultant model performance on medical tasks. 
The results, summarized in Table~\ref{tab: exp_general}, indicate that while medical pre-trained weights offer a slight advantage, the tunable prompts significantly close the gap between non-medical and medical pre-training sources. 
This result underscores the adapter's ability to effectively leverage learned features, irrespective of the pre-training domain, thereby enhancing the model's generalizability.

\begin{table*}[t]
	\renewcommand\tabcolsep{6.0pt}
	\centering
        \caption{Patient ID Out-Of-Distribution results of visual prompt tuning of non-medical pre-trained vision transformers on 160 patients. Total Params denote the total number of parameters for the ViT-B backbone, prompt tokens, and the task heads.}
   \label{tab: exp_patientid_split}
	\scalebox{0.75}{
		\begin{tabular}{c|c|cc|cc|cc}
			\toprule
		Total Params & 160 Seen & 100 Seen & 60 Unseen & 80 Seen & 80 Unseen & 60 Seen & 100 Unseen \\
			\midrule     
            1.39X & \bf 44.87      & \bf 44.56      & \bf 44.97       & \bf 44.23     & \bf 45.01       & \bf 44.19     & \bf 44.85        \\
            1.17X & 43.95      & 43.88      & 43.92       & 43.78     & 44.05       & 43.65     & 43.87        \\
            1.10X & 43.59      & 43.45      & 43.37       & 43.38     & 43.55       & 43.23     & 43.02        \\
            1.05X & 43.21      & 43.12      & 42.85       & 42.96     & 42.89       & 42.87     & 42.68        \\
            1.02X & 42.87      & 42.68      & 42.27       & 42.32     & 42.15       & 42.09     & 41.87        \\
            1.01X & 42.35      & 42.09      & 41.55       & 41.63     & 41.46       & 41.35     & 41.16   \\
			\bottomrule
			\end{tabular}}
  \vspace{-0.5em}
\end{table*}

\begin{table}[t]
	\renewcommand\tabcolsep{6.0pt}
	\centering
        \caption{Patient ID Out-Of-Distribution results of visual prompt tuning of non-medical pre-trained vision transformers on 80 patients seen in the training set. Total Params denote the total number of parameters for the backbone encoder ViT-B, prompt tokens, and the task heads.}
   \label{tab: exp_patientid_unseen}
	\scalebox{0.75}{
		\begin{tabular}{c|c|cccc}
			\toprule
		Total & \# Seen & \multicolumn{4}{c}{\# Unseen} \\
            Params & 80 & 80 & 60 & 40 & 20 \\
			\midrule     
            1.39X & \bf 44.23     & \bf 45.01       & \bf 44.98       & \bf 45.03       & \bf 45.06       \\
            1.17X & 43.78     & 44.05       & 44.08       & 44.06       & 44.09       \\
            1.10X & 43.38     & 43.55       & 43.52       & 43.57       & 43.59       \\
            1.05X & 42.96     & 42.89       & 42.87       & 42.85       & 42.88       \\
            1.02X & 42.32     & 42.15       & 42.13       & 42.11       & 42.12       \\
            1.01X & 41.63     & 41.46       & 41.43       & 41.42       & 41.45       \\
			\bottomrule
			\end{tabular}}
\end{table}

\begin{table}[t]
	\renewcommand\tabcolsep{6.0pt}
	\centering
        \caption{Patient ID Out-Of-Distribution results of visual prompt tuning of non-medical pre-trained vision transformers on 20 patients unseen in the training set. Total Params denote the total number of parameters for the backbone encoder ViT-B, prompt tokens, and the task heads.}
   \label{tab: exp_patientid_seen}
	\scalebox{0.75}{
		\begin{tabular}{c|ccccc}
			\toprule
		Total & \multicolumn{5}{c}{\# Seen} \\
            Params & 140 & 120 & 100 & 80 & 60 \\
			\midrule     
            1.39X & \bf 43.76 & \bf 44.35 & \bf 44.57 & \bf 45.06 & \bf 44.25 \\
            1.17X & 43.33 & 43.69 & 43.89 & 44.09 & 43.68 \\
            1.10X & 43.01 & 43.25 & 43.46 & 43.59 & 43.27 \\
            1.05X & 42.65 & 42.79 & 43.15 & 42.88 & 42.85 \\
            1.02X & 41.86 & 42.09 & 42.65 & 42.12 & 42.05 \\
            1.01X & 41.15 & 41.39 & 42.05 & 41.45 & 41.31 \\
			\bottomrule
			\end{tabular}}
  \vspace{-0.5em}
\end{table}

\subsection{Patient ID Out-of-Distribution}

To assess the impact of patient ID out-of-distribution on medical visual adaptation, we conducted experiments using visual prompt tuning~\cite{jia2022vpt} on 160 infected patients in LHNCBC Malaria~\cite{Lhncbc} dataset with different patient ID splits, including seen and unseen distributions. 
Specifically, we utilize three settings for a comprehensive evaluation:
1) use the same total number of seen and unseen patients, 
2) use the same number of seen patients and a varied number of unseen patients from $\{80,60,40,20\}$;
3) use the same number of unseen patients and a varied number of seen patients from $\{140,120,100,80,60\}$.
For the first setting, we vary the number of seen patients from $\{160, 100, 80, 60\}$ and the number of unseen patients from $\{0, 60, 80, 100\}$.
The results in Tables~\ref{tab: exp_patientid_split}, \ref{tab: exp_patientid_unseen}, and \ref{tab: exp_patientid_seen}, demonstrate the robustness of prompt tuning to patient ID variations. 
We can observe that with the increase in the number of tunable parameters from 1.01X to 1,39X, the model’s performance improves regarding all settings.
When using the same number of unseen patients for evaluation, increasing the number of seen patients from 60 to 80 enhances the results. 
However, increasing 80 to 140 deteriorates the results, which might be caused by the overfitting of the training sets.
While performance naturally dips in unseen patient ID scenarios, the adapter maintains a commendable level of accuracy, suggesting its potential to handle real-world variations and generalize across patient cohorts effectively.

\section{Conclusion}

In this work, we introduced Med-VTAB, a comprehensive benchmark for medical visual task adaptation, addressing a significant gap in the evaluation of pre-trained Vision Transformers (ViTs) across diverse medical imaging modalities and tasks. 
Through the deployment of Med-VTAB, we have explored the landscape of medical visual adaptation with an emphasis on organ diversity, modality diversity, and adaptation diversity, encompassing a vast dataset of 1.68 million medical images.
Our methodological contribution, the Gated Mixture-of-Experts Adapter (GMoE-Adapter), represents a novel approach in leveraging both medical and general pre-trained weights for enhancing the performance of ViTs on medical visual tasks. 
The experimental analysis provided deep insights into three critical aspects of medical visual adaptation: the scaling law of medical prompt tuning concerning tunable parameters, the generalizability of adaptation using both non-medical and medical pre-train weights, and the impact of patient ID out-of-distribution scenarios on model performance. 
Our findings underscore the importance of scalable, generalizable, and robust adaptation mechanisms in medical imaging, where variability across patients and tasks is immense.

\noindent\textbf{Broader Impact.}
Our work on Med-VTAB and the GMoE-Adapter not only advances the state of the art in medical visual task adaptation but also opens up new avenues for research. 
Our contributions pave the way for more personalized, accurate, and accessible diagnostic tools, ultimately contributing to better healthcare outcomes through the power of artificial intelligence.
While our work has the potential to improve medical imaging analysis significantly and show robustness against out-of-distribution data, it also raises important ethical considerations. 
The adaptation of AI models in healthcare must be approached with caution to ensure patient privacy, data security, and bias mitigation.

%
%
\bibliographystyle{splncs04}
\bibliography{reference}

\newpage

\appendix
\section*{Appendix}

In this appendix, we further provide the following materials:
\begin{itemize}
    \item additional details about datasets in Section~\ref{sec:dataset};
    \item additional discussions on the scaling law across diverse organs in Section~\ref{sec:scaling};
    \item additional analysis about patient ID Out-of-Distribution in Section~\ref{sec:ood}.
\end{itemize}

\begin{table*}[t]
	\renewcommand\tabcolsep{3.0pt}
	\centering
        \caption{Details about dataset source, topic, disease, modality, and number of samples in our benchmark.}
   \label{tab: dataset}
	\scalebox{0.8}{
		\begin{tabular}{lcccc}
			\toprule
			\bf Dataset & \bf Topic & \bf Disease & \bf Modality & \bf \# Data \\
			\midrule
HyperKvasir~\cite{borgli2020HyperKvasir}                     & Polyp                    & Pathological                                    & Color images      & 110079             \\
MESAD Prostatectomy~\cite{Bawa2021ESAD}                  & Prostatectomy & Prostatectomy Surgey & Color images      & 29454              \\
AMLC~\cite{matek2019human}           & Cell                     & Peripheral blood smears                         & Color images      & 18365              \\
APTOS~\cite{aptos}                   & Eye                      & Severity of diabetic   retinopathy              & Color images      & 3662               \\
ISIC~\cite{isic}          & Skin                     & Skin Cancer                                     & Color images      & 25331              \\
Kvasir~\cite{Kvasirv2}                     & Gastrointestinal         & Gastrointestinal cancer                         & Color images      & 6000               \\
LHNCBC~\cite{Lhncbc}                & Cells                    & Malaria                                         & Color images      & 27560              \\
MLLBone~\cite{Matek2021highly}               & Cells                    & Blood cell                                      & Color images      & 171374             \\
EyePACS~\cite{eyepacs}                         & Eye                      & Diabetic Retinopathy                            & Color images      & 88702              \\
Vindr~\cite{Nguyen2022vindr}                           & Lung                     & Abnormal Disease                                & X-ray             & 18000              \\
CBIS~\cite{lee2017cbis}                       & Breast                   & Breast Cancer                                   & X-ray             & 10239              \\
COVIDx~\cite{wang2020covidx}                          & Lung                     & Covid-19                                        & X-ray             & 194922             \\
SYMH~\cite{symh} & Shoulder                 & Shoulder X-ray                                  & X-ray             & 1049               \\
RSNA Bone~\cite{Halabi2019TheRP}                 & Bone                     & Bone age                                        & X-ray             & 12611              \\
CheXpert~\cite{Irvin2019chexpert}                        & Chest                    & Radiograph                                      & X-ray             & 224,316            \\
RSNA~\cite{shih2019rsna}                            & Lung                     & Abnormal Disease                                & X-ray             & 29,684             \\
Heidelberg   OCT~\cite{Kermany2018LargeDO}                & Eye                      & OCT Imaging                                     & OCT               & 84495              \\
CC-CCII~\cite{zhang2020clinically}                         & Chest                    & COVID-19 Pneumonia                              & CT                & 617,775            \\
Mosmed~\cite{Morozov2020MosMedDataCC}                           & Chest                    & COVID-19 Pneumonia                              & CT                & 1110               \\
COVID-C~\cite{RAHIMZADEH2021102588}                     & Chest                    & COVID-19 Pneumonia                              & CT                & 349                \\
RICORD~\cite{tsai2021ricord}                          & Chest                    & COVID-19 Pneumonia                              & CT                & 120                \\
PPMI~\cite{Kenneth2011ppmi}                            & Brain                    & Parkinson                                       & MRI               & 480                \\
Brain-Tumor~\cite{braintumor}                     & Brain                    & Brain Tumor                                     & MRI               & 7023              \\
			\bottomrule
			\end{tabular}}
  \vspace{-0.5em}
\end{table*}

\section{Dataset Details}\label{sec:dataset}

In this section, we provide detailed information on the datasets incorporated into the Med-VTAB benchmark. These datasets span across different imaging modalities such as Color Images, X-ray, Optical Coherence Tomography (OCT), Computed Tomography (CT), and Magnetic Resonance Imaging (MRI), reflecting the diversity and complexity of medical visual tasks, as shown in Table~\ref{tab: dataset}.
The comprehensive details including the dataset names, primary medical applications, number of images, unique characteristics, and data sizes are detailed as follows.

\subsection{Color Images}

\begin{itemize}
    \item \textbf{HyperKvasir}~\cite{borgli2020HyperKvasir}: Polyp, Color images. The dataset contains 110,079 images where it captures anatomical landmarks and pathological/normal findings across 23 different classes. \footnote{\url{https://datasets.simula.no/downloads/hyper-kvasir/hyper-kvasir.zip}}.
    \item \textbf{MESAD Prostatectomy}~\cite{Bawa2021ESAD}: Prostatectomy Procedures, Action classification in Prostatectomy Surgery. The dataset includes 29,454 color images with 21 different action classes in the dataset. \footnote{\url{https://saras-esad.grand-challenge.org/download/}}. 
    \item \textbf{AMLC}~\cite{matek2019human}: Cell, Peripheral blood smears, Color images. The dataset consists of 18,365 images across 15 different morphological classes. \footnote{\url{https://www.kaggle.com/datasets/rashasalim/blood-smear-images-for-aml-diagnosis/download?datasetVersionNumber=1}}. 
    \item \textbf{APTOS}~\cite{aptos}: Eye, Severity of diabetic retinopathy, color images. The dataset contains 3,662 images, rated on a scale of 0 to 4 for the severity of diabetic retinopathy across 5 different classes. \footnote{\url{https://www.kaggle.com/competitions/aptos2019-blindness-detection/data}}. 
    \item \textbf{ISIC}~\cite{isic}: Skin Cancer, Color images. This dataset includes 25,331 images for the classification of dermoscopic images across 9 different diagnostic categories. \footnote{\url{https://www.kaggle.com/datasets/andrewmvd/isic-2019/download?datasetVersionNumber=1}}. 
    \item \textbf{Kvasir}~\cite{Kvasirv2}: Gastrointestinal cancer, Color images. The dataset includes 6,000 images showing 8 different classes of findings in the GI tract. \footnote{\url{https://www.kaggle.com/datasets/plhalvorsen/kvasir-v2-a-gastrointestinal-tract-dataset/download?datasetVersionNumber=1}}. 
    \item \textbf{LHNCBC Malaria}~\cite{Lhncbc}: Cells, Malaria, Color images. The dataset contains 27,560 images with 12 different classes of annotations from 160 patients. \footnote{\url{https://lhncbc.nlm.nih.gov/LHC-research/LHC-projects/image-processing/malaria-datasheet.html}}. 
    \item \textbf{MLLBone}~\cite{Matek2021highly}: Cells, Blood cell, Color images. This dataset features 171,374 images across 21 different classes of tissues from 945 patients. \footnote{\url{https://wiki.cancerimagingarchive.net/pages/viewpage.action?pageId=101941770\#1019417705cfb4244105f44a68817f524de9fc7e2}}.
    \item \textbf{EyePACS}~\cite{eyepacs}: Eye, Diabetic Retinopathy, Color images. The dataset includes 88,702 images categorized into 5 classes based on the severity of Diabetic Retinopathy (DR). \footnote{\url{https://www.kaggle.com/competitions/diabetic-retinopathy-detection/data}}. 
\end{itemize}

\subsection{X-ray}

\begin{itemize}
    \item \textbf{Vindr}~\cite{Nguyen2022vindr}: Abnormal Disease. This dataset contains 18,000 CXR scans with 14 critical findings. \footnote{\url{https://www.kaggle.com/competitions/vinbigdata-chest-xray-abnormalities-detection/data}} 
    \item \textbf{CBIS-DDSM}~\cite{lee2017cbis}: Breast Cancer. This dataset contains 10,239 images for breast cancer screening, including conditions such as normal, benign, and malignant. \footnote{\url{https://wiki.cancerimagingarchive.net/download/attachments/22516629/CBIS-DDSM-All-doiJNLP-zzWs5zfZ.tcia?version=1\&modificationDate=1534787024127\&api=v2}}
    \item \textbf{COVIDx}~\cite{wang2020covidx}: Lung Covid-19. This dataset includes 194,922 images for COVID-19 detection, categorized into 4 classes.\footnote{\url{https://www.kaggle.com/datasets/andyczhao/covidx-cxr2/download?datasetVersionNumber=9}}
    \item \textbf{SYMH}~\cite{symh}: Shoulder. This dataset provides 1,049 shoulder X-ray images across 4 categories.~\footnote{\url{https://www.kaggle.com/datasets/dryari5/shoulder-xray-classification}}
    \item \textbf{RSNA Bone}~\cite{Halabi2019TheRP}: Bone age. This dataset features 12,611 images for bone age assessment, spanning 228 age classes.~\footnote{\url{https://www.kaggle.com/datasets/kmader/rsna-bone-age/download?datasetVersionNumber=2}}
    \item \textbf{CheXpert}~\cite{Irvin2019chexpert}: Chest radiographs. This dataset includes 224,316 chest radiographs with 5 categories (atelectasis, cardiomegaly, consolidation, edema, and pleural effusion) from 65,240 patients. \footnote{\url{https://stanfordmlgroup.github.io/competitions/chexpert/}}
    \item \textbf{RSNA}~\cite{shih2019rsna}: This dataset contains 29,684 chest radiographs with 2 classes (normal and pneumothorax positive). \footnote{\url{https://pubmed.ncbi.nlm.nih.gov/33937785/}}.
\end{itemize}

\subsection{Optical Coherence Tomography (OCT)}

\textbf{Heidelberg OCT}~\cite{Kermany2018LargeDO}: contains 84,495 OCT images across 4 categories (CNV, DME, DRUSEN, and NORMAL) related to eye diseases. \footnote{\url{https://data.mendeley.com/public-files/datasets/rscbjbr9sj/files/5699a1d8-d1b6-45db-bb92-b61051445347/file_downloaded}}.

\subsection{Computed Tomography (CT)}

\begin{itemize}
    \item \textbf{CC-CCII}~\cite{zhang2020clinically}: focuses on COVID-19 pneumonia with a large dataset of 617,775 CT images.\footnote{\url{http://ncov-ai.big.ac.cn/download?lang=en}}
    \item \textbf{Mosmed}~\cite{Morozov2020MosMedDataCC}: documents COVID-19 pneumonia cases in a collection of 1,110 CT scans.\footnote{\url{https://academictorrents.com/details/f2175c4676e041ea65568bb70c2bcd15c7325fd2}}
    \item \textbf{COVID-C}~\cite{RAHIMZADEH2021102588}: contains 349 CT images for COVID-19 pneumonia detection.\footnote{\url{https://github.com/mr7495/COVID-CTset}}
    \item \textbf{RICORD}~\cite{tsai2021ricord}: offers a dataset of 120 CT images also for COVID-19 pneumonia analysis.\footnote{\url{https://wiki.cancerimagingarchive.net/pages/viewpage.action?pageId=80969771}}
\end{itemize}

\subsection{Magnetic Resonance Imaging (MRI)}

\begin{itemize}
    \item \textbf{PPMI}~\cite{Kenneth2011ppmi}: A dataset containing 480 MRI scans related to Parkinson's disease.\footnote{\url{https://www.ppmi-info.org/access-data-specimens/download-data}}
    \item \textbf{Brain-Tumor}~\cite{braintumor}: consists of 7,023 MRI images for brain tumor detection and segmentation.\footnote{\url{https://www.kaggle.com/datasets/masoudnickparvar/brain-tumor-mri-dataset}}
\end{itemize}

\section{Additional Analysis on Scaling Law}\label{sec:scaling}

Table 5 presents a detailed examination of the scaling law of medical visual prompt tuning with DINO v2~\cite{oquab2023dinov2} pre-trained vision transformers on color images. We observe that the performance improvement is directly correlated with the increase in the total number of parameters (Total Params) which encompasses the backbone encoder ViT-B~\cite{dosovitskiy2021an}, prompt tokens, and the task heads.
An analysis across different organs and medical conditions—such as cell, eye, skin, and polyp—reveals quantitative differences in the adaptation efficiency of tunable parameters. 
For example, the HyperKvasir Polyp~\cite{borgli2020HyperKvasir} dataset demonstrates substantial performance improvements as Total Params increases, whereas the AMLC~\cite{matek2019human} Cell dataset shows relatively modest gains. 

Based on the observed data, we propose the following suggestions for tuning parameters tailored to each organ type. 
For polyp detection tasks (HyperKvasir~\cite{borgli2020HyperKvasir} and Kvasir~\cite{Kvasirv2} Polyp), a higher number of tunable parameters (>1.1X) can significantly boost performance, likely due to the complexity and variability within the image data. 
Cell-related tasks (AMLC~\cite{matek2019human} and MLLBone~\cite{Matek2021highly} Cell) may require a moderate increase in parameters, as excessive tuning does not correspond to a significant performance leap, suggesting a point of diminishing returns. 
Eye conditions (APTOS~\cite{aptos} and EyePACS~\cite{eyepacs}) show a favorable response to parameter scaling, indicating that a careful increment in parameters could be beneficial, particularly for nuanced features in retinal images.
For skin lesion analysis (ISIC Skin~\cite{isic}), a moderate to high parameter scaling appears to offer the best performance gains, potentially due to the visual similarity among different lesion types which requires more complex model tuning to distinguish.

\section{Additional Analysis on Patient ID Out-of-Distribution}\label{sec:ood}

To delve deeper into the influence of Patient ID Out-of-Distribution (OOD) on medical visual adaptation, we extended our experiments with visual prompt tuning as proposed by Jia et al.\cite{jia2022vpt} using the LHNCBC Malaria dataset\cite{Lhncbc}, focusing on 160 infected patients and analyzing the effects under various patient ID splits. 
Our analysis not only considers the variation in the number of seen versus unseen patients but also investigates the impact of tuning parameter volume on the OOD challenge. 
The experimental setup and findings are further enriched by exploring nuances that deviate from expected performance trends against unseen patient data.

\textbf{Impact of Tunable Parameters.}
The volume of parameters tuned during adaptation plays a crucial role in the model's ability to generalize to unseen patient IDs. 
Our observations align with the intuition that an increase in tunable parameters (ranging from 1.01X to 1.39X) correlates with improved model performance across all settings. 
This enhancement suggests that a larger parameter space provides the model with more flexibility to capture and generalize the diverse patterns present in medical images, thereby improving its robustness to variations in patient IDs.

\textbf{Unseen Patients and Performance Trends.}
Contrary to the straightforward expectation that an increase in the number of unseen patients would linearly degrade model performance, our analysis reveals a more nuanced interaction (refer to Table 8). 
This phenomenon indicates that the relationship between the number of unseen patients and model performance is not linear and might be influenced by the complexity and distribution of the dataset. 
One possible explanation could be that the diversity within the seen patient group provides a rich enough representation of the dataset's variability, allowing the model to maintain or even improve its performance up to a certain threshold of unseen patients.

\textbf{The Balance Between Seen and Unseen Patients.}
Our findings also highlight a delicate balance in the number of seen versus unseen patients (refer to Table 9). 
While increasing the number of seen patients from 60 to 80 improves performance, further increases up to 140 lead to a decrease in performance.
This trend suggests a potential overfitting to the training set, where the model becomes too specialized in the representations of the seen patients, diminishing its ability to generalize to new, unseen patients.
It underscores the importance of selecting an optimal balance of seen and unseen patients to avoid overfitting while maximizing the model's generalization capabilities.

In summary, our additional analysis underscores the complex dynamics at play in adapting medical visual models to handle patient ID out-of-distribution effectively. 
The impact of tunable parameters, the nuanced relationship between unseen patient numbers and performance, and the potential for overfitting with an excessive focus on seen patients, all contribute to our understanding of model adaptation in the face of patient variability. 
These insights not only bolster the robustness and applicability of visual prompt tuning in medical imaging but also pave the way for future research to explore more sophisticated adaptation techniques that can navigate the intricacies of real-world medical data distributions.

\end{document}